\documentclass{bmvc2k}
\usepackage{enumerate}
\newcommand{\RNum}[1]{\uppercase\expandafter{\romannumeral #1\relax}}
\usepackage{amsmath}
\usepackage{stfloats}
\usepackage{float}
\usepackage{threeparttable}
\usepackage{subfigure}
\usepackage{lineno,hyperref}
\usepackage{graphicx}
\usepackage{amsmath}
\usepackage{amssymb}
\usepackage{booktabs}
\usepackage{enumerate}
\usepackage{algorithm} 
\usepackage{algpseudocode}

\usepackage{multirow}
\usepackage{subfigure}
\usepackage{stfloats}
\title{Spatio-temporal Tendency Reasoning \\
for Human Body Pose and Shape Estimation from Videos}

 \addauthor{Boyang Zhang}{boyangchang@foxmail.com}{1}
 \addauthor{Suping Wu{*}}{pswuu@nxu.edu.cn}{1}
 \addauthor{Hu Cao}{caohu19980219@gmail.com}{1}
 \addauthor{Kehua Ma}{919056348@qq.com}{1}
 \addauthor{Pan Li}{18235448104@163.com}{1}
 \addauthor{Lei Lin}{2409937088@qq.com}{1}

 \addinstitution{
  School of information Engineering\\
  Ningxia University\\
  Ningxia, CHN\\
 }
\runninghead{Zhang et al.}{Spatio-temporal Tendency Reasoning 
for Pose and Shape}

\begin{document}

\maketitle
% \vspace{-0.4cm}
\begin{abstract}

In this paper, we present a spatio-temporal tendency reasoning (STR) network for recovering human body pose and shape from videos. Previous approaches have focused on how to extend 3D human datasets and temporal-based learning to promote accuracy and temporal smoothing. Different from them, our STR aims to learn accurate and natural motion sequences in an unconstrained environment through temporal and spatial tendency and to fully excavate the spatio-temporal features of existing video data. To this end, our STR learns the representation of features in the temporal and spatial dimensions respectively, to concentrate on a more robust representation of spatio-temporal features. More specifically, for efficient temporal modeling, we first propose a temporal tendency reasoning (TTR) module. TTR constructs a time-dimensional hierarchical residual connection representation within a video sequence to effectively reason temporal sequences' tendencies and retain effective dissemination of human information. Meanwhile, for enhancing the spatial representation, we design a spatial tendency enhancing (STE) module to further learns to excite spatially time-frequency domain sensitive features in human motion information representations. Finally, we introduce integration strategies to integrate and refine the spatio-temporal feature representations. Extensive experimental findings on large-scale publically available datasets reveal that our STR remains competitive with the state-of-the-art on three datasets. Our code are available at \url{https://github.com/Changboyang/STR.git}.
\end{abstract}
% \vspace{-0.4cm}
%-------------------------------------------------------------------------
\section{Introduction and Related Work}
\label{sec:intro}
The basic goal of 3D human body pose and shape estimation (a.k.a., 3D human motion estimation) in video aims to estimate 3D human pose and shape from motion videos, which have a wide range of computer vision applications. Existing methods are mainly based on parametric models such as SMPL \cite{loper2015smpl} and SCAPE \cite{anguelov2005scape} to represent the human body. Depending on the input, we can roughly divide existing methods into two categories: Image-based methods \cite{kanazawa2018end, kolotouros2019learning, xu20213d} and video-based methods \cite{kanazawa2019learning, kocabas2020vibe, xu20213d, choi2021beyond}. The latter not only needs to ensure the single-frame image reconstruction effect but also to recover a time-smoothed human video. So it is more complicated than single-frame image reconstruction. Kanazawa et al. \cite{kanazawa2019learning} encode temporal features via 1D convolution. Although this method obtains smoother human body sequences, the insufficient modeling of spatial features leads to lower accuracy of estimated human poses. 
Based on this, Sun et al. \cite{sun2019human} propose a skeleton decoupling-based paradigm to improve spatial accuracy. Although this approach improves spatial accuracy, it neglects the grasp and reasoning of spatio-temporal feature tendency and fails to balance temporal smoothness and spatial accuracy. Kocabas et al. \cite{kocabas2020vibe} train an adversarial learning network and use AMASS \cite{mahmood2019amass} to discriminate between real human motion and human motion generated by the temporal human body pose and shape regression network. Later, Choi et al. \cite{choi2021beyond} abandon the strong dependence on the current static frame and propose a mesh recovery system for PoseForecast that effectively pays attention to temporal information.
\begin{figure*}
\centering
\includegraphics[width=\textwidth]{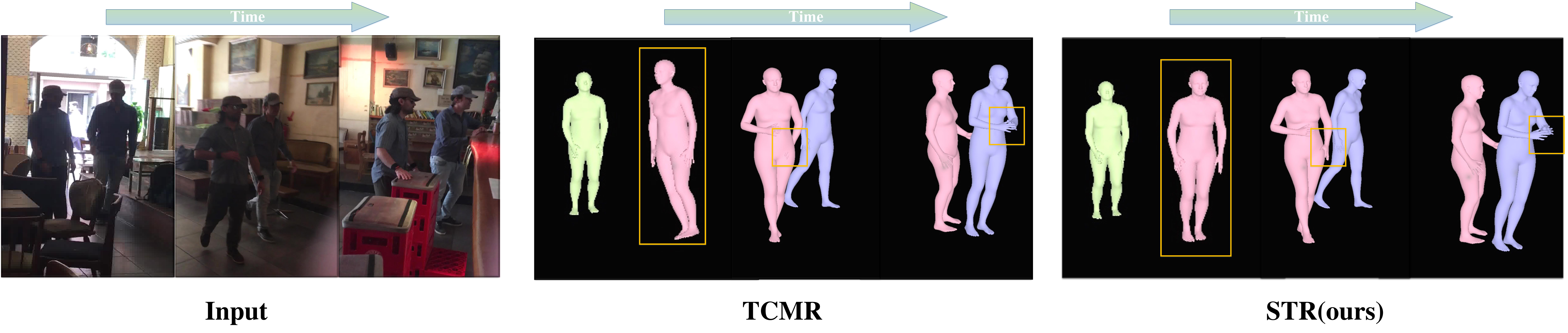}
\vspace{-0.5cm}
\caption{From left to right are the input video sequence, the reconstructed human sequence of TCMR \cite{choi2021beyond}, and the reconstructed human sequence of STR. STR demonstrated more realistic and smoother human action in extreme light illumination than the SOTA method TCMR.} \label{fig:image1}
% \vspace{-0.35cm}
\label{fig1}
\end{figure*}

However, such methods for estimating human pose and shape from videos still have limited performance on some challenging problems. As shown in Figure \ref{fig:image1}, when capturing motion images in an unconstrained environment (low natural illumination and blurred human motion), it leads to poor model parameter estimation and thus reconstructs an unreasonable human body. Although some approaches \cite{Joo2021ExemplarFF, Sengupta2020SyntheticTF, kocabas2020vibe} attempt to improve performance by adding external data resources, these methods do not take full advantage of the potential information in the underlying data. Meanwhile, some methods \cite{choi2021beyond, luo20203d} are inherently limited to modeling video temporal relationships. While these methods improve the temporal consistency of human pose estimation in video, they lack spatial understanding and reasoning capabilities, leading to biased predictions.
% In addition, the process of modeling spatial features and temporal consistency is mostly learned on a separate basis. This is likely to fall into a sub-optimal state, resulting in a lack of connection between time and space. 
In general, during human movement, the current motion depends on the state of the previous motion and influences the subsequent motion sequences. However, when there are problems with extreme illumination or motion blur in the video, the current motion does not effectively obtain the state of the previous motion and can negatively affect future motion. Since human motion has a similar development tendency, we call the above problem tendency reasoning. We find that temporal tendency reasoning helps to explore long dependencies between frames and to obtain information from frames with a larger temporal range. The enhancement of spatial tendency helps the network to focus more on human-related features in unconstrained scenes and mitigate background effects, thus estimating the pose accurately for each frame.

Based on the above observations and problems, we offer a spatio-temporal tendency reasoning (STR) approach for estimating human body pose and shape from videos. Our STR reason temporal and spatial tendencies separately and use integration strategies to aggregate them with each other. In particular, in the temporal tendency reasoning (TTR) module, in order to preserve the efficient dissemination of temporal tendency in motion sequences, we partition spatial features and the corresponding recurrent layers into hierarchical subsets, in which the network reasons the temporal tendency in each subset in an incremental manner.
Subsets of different layers are then concatenated together by a residual structure to reason the tendency of the whole motion sequence.
For the modeling of spatial tendency, we employ the means of spatial tendency enhancement to learn human movement. Existing work utilizes VAE \cite{yan2018mt, luo20203d} and optical flow methods \cite{zou2021eventhpe} to learn human action representations. Unlike them, our spatial tendency enhancing (STE) module models the human motion representation by calculating the difference in motion between adjacent frames. In STE, we perform time-domain spatial enhancement and frequency-domain spatial enhancement separately. Both employ motion representations to adaptively generate weights. These weights can be used to excite spatial-sensitive features in the time domain and high-frequency motion features in the frequency domain, allowing the network to uniformly learn human body spatial features and motion features. Furthermore, we introduce integration strategies to refine and integrate human features through self-integration strategy and cross-integration strategy.
% We employ feature-level difference representations to characterize human motion. These motion features are then used to adaptively generate the weights that can be dictated to excite spatially sensitive features, allowing the human motion information to be adequately learned by the network. In the meantime, we consider utilizing information from the frequency domain to enhance human body feature information. Furthermore, we employ the self-fusion strategy to improve the representation of human motion features by repeatedly using spatio-temporal feature information.

Our main contributions to this work are outlined below:
\begin{itemize}
\item We propose a spatio-temporal tendency reasoning for human body pose and shape estimation from videos, which can alleviate the problem of human reconstruction in unconstrained scenes.
\item We design a temporal tendency reasoning module and a spatial tendency enhancement module, respectively, to facilitate the effective propagation of motion information over long-distance frames and to stimulate spatially sensitive features. We also propose integration strategies module to enhance the integrated representation of different features.
\item Experimentally, both the quantitative and qualitative results of our method show the effectiveness of the proposed method on widely evaluated benchmark datasets.
\end{itemize}
\begin{figure*}
\centering
\includegraphics[width=1.0\textwidth]{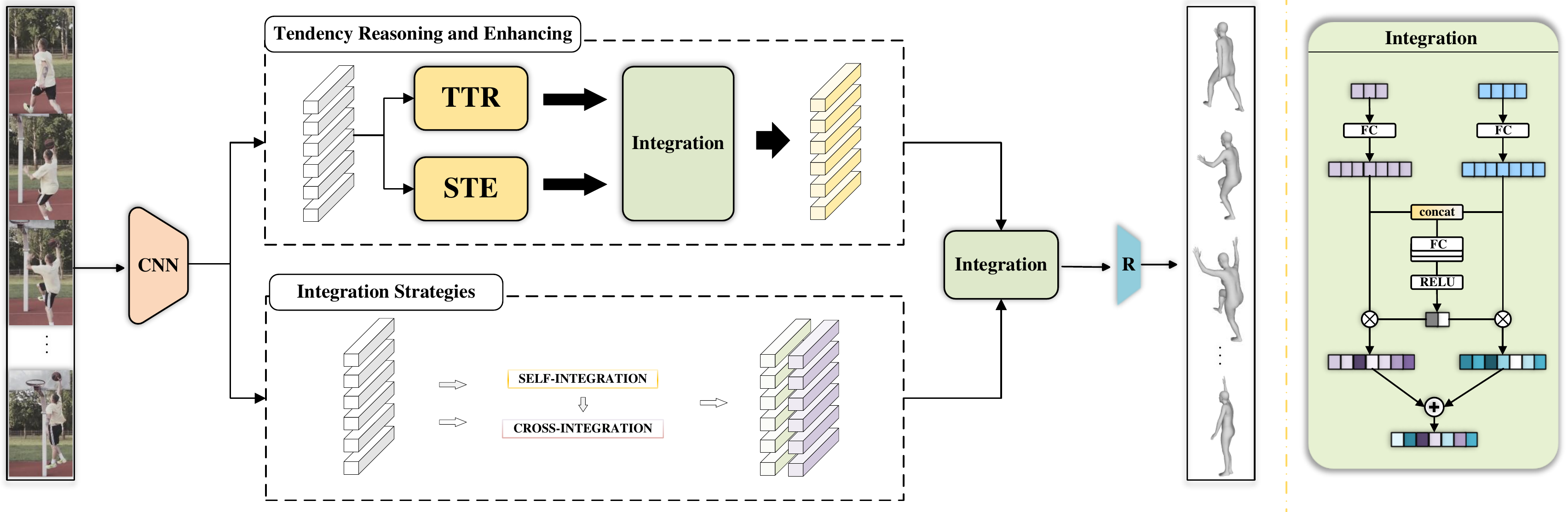}
\caption{An overview of our framework. Given a video sequence, the aim of our method is to reconstruct the corresponding human sequence. Our method consists of two modules, a tendency reasoning enhancing module and an integration strategy module. The tendency reasoning enhancing module consists of a temporal tendency reasoning and a spatial tendency enhancing module. The integration strategies consist of a self-integration strategy and a cross-integration strategy. }\label{fig:image2}
\label{fig2}
% \vspace{-0.2cm}
\end{figure*}
% \vspace{-0.4cm}

\section{Approach}
Figure \ref{fig:image2} shows the overall pipeline of our STR. Given an input video $V=\{I_{t}\}_{t=1}^{T}$ of length $T$, we utilize the ResNet-50 \cite{he2016deep} to extract feature vectors $F=\{f_{i}\}_{i=1}^{T}\in\mathbb{R}^{T \times 2048}$ of each frame. Next, $F$ passes through the TTR and STE modules to reason the temporal tendency and enhance the spatial tendency respectively and fuses the results of the two modules through the integration network to output the enhanced tendency features. Meanwhile, $F$ is fed into self-integration and cross-integration to output two enhanced spatio-temporal features. Ultimately, the outputs of these modules are fused through an integration network and the results are used to regress the SMPL parameters.
\subsection{SMPL Model}
We employ the SMPL statistical model to characterize humans.
% The skinned multi-person linear model (SMPL) \cite{loper2015smpl} is one of the broadly available statistical human models. The pose parameter $\theta\in\mathbb{R}^{3J}$ represents the axis angle representation for relative rotation of the $J$ skeletal joints in the kinematic tree relative to their parent joints, where $J$ = 24. The shape parameter $\beta\in\mathbb{R}^{10}$ indicates the PCA basis coefficients to express the body shape.
SMPL defines a function $M(\theta, \beta)$, which takes a set of pose parameters $\theta\in\mathbb{R}^{3\times J}$ of the $J$ skeletal joints and shape parameters $\beta\in\mathbb{R}^{10}$ as input, and outputs a full-body triangulated mesh $M\in\mathbb{R}^{N\times3}$ with $N$ = 6890 vertices. The model transforms the mesh vertices to the body joints $J$ by a mapping, here $J = W\cdot{M}$, where $W$ is a pre-trained linear regressor.

\subsection{Temporal Tendency Reasoning}
\begin{figure}
\centering
\includegraphics[width=1\textwidth]{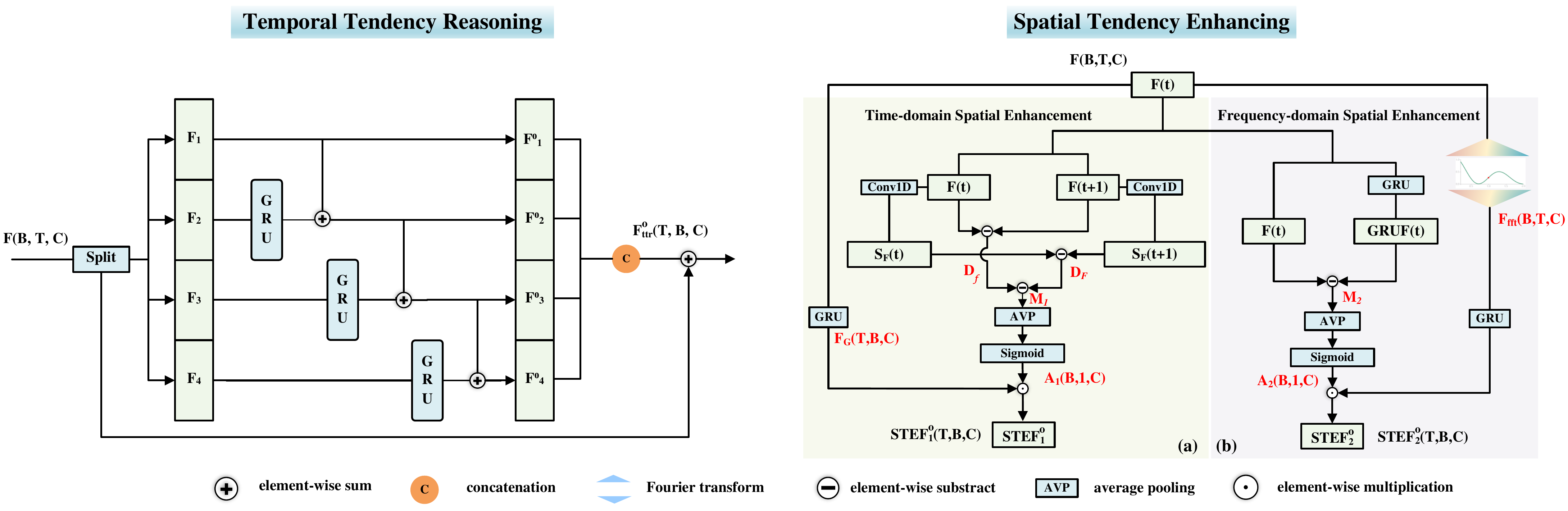}
\caption{Illustration of the temporal tendency reasoning module (left) and spatial tendency enhancing (right). TTR module inputs feature $F$ and outputs the reasoned spatio-temporal feature $F^o_{ttr}$. STE module inputs feature $F$ and then passes through time-domain spatial enhancement (a) and frequency-domain spatial enhancement (b) respectively, finally outputting two different enhanced spatio-temporal features $STEF_{1}^{o}$, $STEF_{2}^{o}$.}\label{fig:image3}
\label{fig4}
\end{figure}

As shown in Figure \ref{fig:image3}, we split $F$ evenly into 4 sub-fragments in the temporal dimension to construct 4 sub-branches, where each fragment is shaped as $B\times \frac{T}{4}\times C$. More specifically, the $F_1$ sub-fragment does not undergo any operation, our TTR takes the other three sub-fragments from $F_2$ to $F_4$ as the inputs to three identical GRU \cite{dey2017gate} to learn temporal representation. To further reason temporal tendency, then for the four branches, our TTR adopts a hierarchical cascade architecture to successively fuse the results of the two adjacent branches and transmits progressively them to the next branch to generate new $F_2$, $F_3$ and $F_4$. We formulate this process as follows,

\begin{equation}
\begin{array}{lr}
F_{i}^{o}={F}_{i}, & i=1, \\\vspace{0.5ex}

F_{i}^{o}= GRU(F_{i}^{o}) + F_{i-1}^{o}, & i=2,3,4 \\
\end{array}
\end{equation}
where $F_i^o\in\mathbb{R}^{B\times \frac{T}{4}\times C}$ is the output of i-th sub-fragment. For $F_1$, we do not temporally encode it to maximize the preservation of spatial features. And for $F_2$, we hope to supplement the current features from the relevant spatial features in $F_1$. Under extreme lighting conditions, fragments interact in this form to facilitate the efficient propagation of invisible information over long-distance frames.
In TTR, different sub-fragments focus on different temporal tendencies in a video. 
TTR can aggregate temporal tendency across multiple fragments to reason temporal tendency across whole motion sequences. This not only explores long-term dependencies between fragments but also captures information from long-distance frames.
% For example, the output of the first sub-fragment is $F_0$ which can only model the temporal representation of that sub-fragment, in other words, the network can only reason about the temporal tendency of the first subfragment period.
% TTR aggregates subsequently different sub-fragments. The output of the last sub-fragment $F_4$ can reason the temporal tendency of the entire motion sequence. 
Eventually, each sub-branch is concatenated and then added to the original feature $F$ to integrate the spatio-temporal feature representation. 
\begin{equation}
\begin{array}{lr}
F_{ttr}^{o}=F + concat(F_{1}^{o}, F_{2}^{o}, F_{3}^{o}, F_{4}^{o}), \vspace{0.6ex}\\
\end{array}
\end{equation}
Where $F_{ttr}^{o}\in\mathbb{R}^{T\times B\times C}$ is the output of the TTR module.
In this way, the temporal modeling of the entire video sequence is transformed into temporal tendency reasoning, i.e., the temporal tendency of different sub-branches are combined hierarchically to form a complete temporal tendency. This temporal tendency reasoning is more conducive to the network's learning of long time sequences.
% $F_{ttr}^{o}$ carries the temporal tendency of the entire video that the tendency can be used to estimate the action of future frames. 
% In this way, the whole video sequence temporal modeling is converted to temporal tendency reasoning, which means that the temporal tendency of distinct sub-branches is combined by the network to constitute a complete temporal tendency. This temporal tendency reasoning is more conducive to the network's learning of temporal representations.

\subsection{Spatial Tendency Enhancing}
When encountering extreme illumination, the network cannot express human-related features well. Motion information is an important clue for understanding human behavior in videos. 
Spatial tendency enhancing(STE) aims to enhance human representation and distinguish human-related features by focusing on motion. We observe that the pixel values of human motion regions change over time larger than background regions. So STE exploits the temporal differences of adjacent frame-level features to focus on motion features while suppressing irrelevant information in the background.
We elaborated on two parts of STE, which are time-domain spatial enhancement and frequency-domain spatial enhancement.
% We apply one-dimensional convolution to the feature space and perform spatial filtering to enhance local expression ability of spatial features and spatially consecutive tendency in the space. 
% We apply 1D convolution to features to enhance the local interaction and expressiveness of spatial features.
As shown on the right of Figure \ref{fig:image3}(a), in time-domain spatial enhancement, we first use 1D convolution on feature $F$ of shape $B\times T\times C$ to learn its time-domain spatial representation $S_{F}$. 

Then our STE iteratively calculates the difference between the features of two adjacent frames to construct a difference sequence $D_f$. For $S_{F}$, our STE models the spatial difference sequence $D_{F}$ in the same way. Finally, the two difference sequences are subtracted to calculate the time-domain spatial representation offset $M_{1}$.
\begin{equation}
\begin{array}{lr}
M_1(t) = D_F(t)-D_f(t),  & 0\textless t \textless T  \vspace{0.5ex}\\
D_F(t)=S_F(t+1)-S_F(t),  
D_f(t)=F(t+1)-F(t) \vspace{0.5ex} & 0\textless t \textless T \vspace{0.5ex}
\end{array}
\end{equation}
where $t$ represents a frame in a sequence of $T$ frames. Then we leverage the global average pooling layer to aggregate temporal information and employ a sigmoid layer to learn the spatial offset weight map $A_1$. 
While learning the offset weight map, we also send the original features to the GRU \cite{dey2017gate} layer to learn temporal representation and obtain spatio-temporal features $F_G$. $F_G$ is multiplied by the spatial motion weight map $A_1$ to obtain time-domain spatio-temporal features $STEF_{1}^{o}$.
\begin{equation}
\begin{array}{lr}
STEF_{1}^{o}(t)=F_G(t)*A_{1}, \\
A_{1}=sigmoid(AVP(M_{1})) , F_G=GRU(F)
\end{array}
\end{equation}
$*$ denotes the channel-wise multiplication. 

The Fourier transform is sensitive to high-frequency features of human motion. For Figure \ref{fig:image3}(b), in frequency-domain spatial enhancement, we first perform the Fast Fourier Transform on $F$ to obtain the frequency domain feature representation and then use the inverse Fast Fourier Transform to convert the frequency domain feature back to the temporal domain feature $F_{fft}$. 
% Due to the change of the observation space, the information in the frequency domain is not similar to the corresponding spatial information in the original video, and some spatial motion features are often highlighted in the frequency domain. This enables the frequency domain features to enhance and complement the feature information in the original video. Through transformation, these features not only retain the original video spatio-temporal information but also obtain additional motion information in the frequency domain.
% The inverse Fourier transform complements the frequency-domain high-frequency information back to the time-domain features and highlights their motion representation. 
We consider that the FFT-IFFT operation can preserve human information and highlight the high-frequency motion features to compensate for the lack of time-domain representation.
Next, we send $F$ and $F_{fft}$ to the GRU \cite{dey2017gate}. 
The Fourier transform is sensitive to overall spatial motion, the $GRU(F)$ and $F$ are overall subtracted to obtain the spatio-temporal offset map $M_{2}$ to model the information of the spatio-temporal difference.
Finally, we apply formulas 4, $M_{2}, F_{fft}$ as input, and output spatio-temporal feature $STEF_{2}^{o}$ with enhanced spatial tendency.
\begin{equation}
\begin{array}{lr}
STEF_{2}^{o}(t)= GRU(F_{fft}(t))*A_{2},\\
A_{2}=sigmoid(AVP(M_{2})) ,\\
M_2(t) = GRU(F(t))-F(t),   & 1\leq t \leq T \\
\end{array}
\end{equation}

STE enhances spatial human motion tendency by focusing on continuous frame differences and overall sequence differences. STE fully considers the properties of the time and frequency domain to model the actual motion features. Meanwhile different from using motion estimation network to learn human motion, STE uniformly learns motion features and spatio-temporal features, which can effectively enhance spatial tendency.

\subsection{Integration Strategies}
Integration strategies are classified into self-integration and cross-integration strategies according to the type of input. The aim is to aggregate the outputs of each component via an integration network.
The integrated network is shown in Figure \ref{fig:image2}. First, the network accepts a set of spatio-temporal features as input, then these features are cascaded and passed through multiple RELU activation functions and FC layers, followed by a Softmax activation function to produce a set of weights. This set of weights is then multiplied by the corresponding features and summed to produce the integration features.

% \begin{figure}[h]
% \centering
% \includegraphics[width=0.55\textwidth, height=0.3\textwidth]{images/FUSION.pdf}
% \caption{Illustration of the fusion network. The fusion network is used to aggregate features.}
% \label{fig5}
% \end{figure}
We first integrate the output of TTR, and STE to obtain the spatio-temporal tendency reasoning feature $F_{STR}$. 
\begin{equation}
\begin{array}{lr}
F_{STR}=Integration(F_{ttr}^{o}, STEF_{1}^{o}, STEF_{2}^{o})
\end{array}
\end{equation}
The implementation procedure of the integration strategies module is described in Algorithm 1. 
\begin{algorithm} 
	\caption{Integration strategies}     % 标题
	 \label{xx}       % 用来引用    
	\begin{algorithmic}[1] % 加上 [1] 表示有序号
	\Require All frame features $F$, number of integration selectable $N$.   %   Requre 等同于 Input
    \Ensure  Enhancing features $\hat{F}$   %   Ensure 等同于 Output
    \State /* Splitting $F$ into $c$ parts */    %   语句
    \State  Get $F^{c_1}, F^{c_2}$ = GRU(SPLIT($F$))    %   语句
    
    \State <PHASE 1: SELF-INTEGRATION PHASE>    %   语句
    \For {$i<N$}
      \State  Get $F_i^{c_1}$ = Integration($F^{c_1}$)
      \State  Get $F_i^{c_2}$ = Integration($F^{c_2}$)
    \EndFor
  \State <PHASE 2: CROSS-INTEGRATION PHASE>
  \State Get $\hat{F}_{SF}$ = Integration($F_i^{c_1}, F_i^{c_2}$)
  \State Get $\hat{F}_{CF}$ = Integration($F^{c_1}, F^{c_2}$)
  \State\Return $\hat{F}_{SF}$, $\hat{F}_{CF}$ % Return 返回语句
  \end{algorithmic} 
\end{algorithm}
The integration strategy first divides the input features according to the temporal dimension to focus on the temporal context. The divided features are then temporally encoded and passed through the self-integration phase, the self-integration process can enhance the expression of the original human features. Finally, the enhanced human features $F_i^{c_1}$ and $F_i^{c_2}$ and the original features $F^{c_1}, F^{c_2}$ are fed into the cross-integration phase that has focused on human information at different times. The integration strategies module outputs $\hat{F}_{SF}$, $\hat{F}_{CF}$.
We finally integrate the $\hat{F}_{SF}, F_{STR}$ and $ \hat{F}_{CF}$ to obtain the final spatio-temporal features $Z$. Meanwhile, we feed $Z$ into the SMPL regressor to regress the SMPL parameters.
\begin{equation}
\begin{array}{lr}
Z=Integration(\hat{F}_{SF}, F_{STR}, \hat{F}_{CF})
\end{array}
\end{equation}

% After obtaining the parameters, we feed them into the function defined by SMPL to build a 3D human mesh with pose and shape. In the testing phase, we just feed $F_{int}$ into the regressor and use the results of the regression as the final human body representation.

\subsection{Loss Function}
L2 loss was applied to 2D and 3D joint coordinates and SMPL parameters during training.
\begin{equation}
L_{\mathcal{G}} = \omega_{3d} \sum_{t=1}^{T}\|X_{t}-\hat{X}_{t}\|_{2} + \omega_{2d} \sum_{t=1}^{T}\|x_{t}-\hat{x}_{t}\|_{2} + \omega_{shape} \|\beta-\hat{\beta}\|_{2}+\omega_{pose} \sum_{t=1}^{T}\|\theta_{t}-\hat{\theta}_{t}\|_{2} \nonumber
\end{equation}
where $X_t$ stands for 3d joints, $x_t$ for 2d joints, $\theta$ and $ \beta$ represent the SMPL parameters. $\omega(\cdot)$ denotes the corresponding loss weights.

\section{Experiments}
% We conducted tests on three widely used data sets to assess the efficacy of the suggested technique. And compare our results to prior cutting-edge approaches based on frames and videos. Finally, we conducted ablation experiments to prove the contribution of each component.
\subsection{Implementation Details}
Following the \cite{choi2021beyond} parameters, we initialize the backbone and regressor with the pre-trained SPIN by setting the length of the input sequence T to 16. With a mini-batch size of 32, the weights are modified via the Adam optimizer. 
In order to save training time and memory, we pre-computed the spatial features from the images through ResNet. The initial learning rate is set at $5\times10^{{-5}}$ and is reduced by a factor of 10 if the accuracy of the 3D pose does not improve after 5 epochs. With a Quadro RTX 6000 GPU, we trained the network for 30 epochs. PyTorch was used to implement the code.

\subsection{Evaluation Datasets and Metrics}
\textbf{Evaluation Datasets.}
The 3DPW\cite{von2018recovering} is a 3D dataset capturing the SMPL human body in a natural scene. The MPI-INF-3DHP\cite{mehta2017monocular} consists of over 1.3 million frames of video of 11 people captured by 14 cameras simultaneously. Human3.6M\cite{ionescu2013human3} is a massive dataset of 3.6 million RGB images of 15 daily activities performed by 11 different professional actors.

\textbf{Evaluation Metrics.}
We calculated the mean error per joint position (MPJPE) and Procrustes-aligned MPJPE (PA-MPJPE) as the main metrics of accuracy. And we measured the Euclidean distance (MPVPE) between the ground truth vertex and the predicted vertex. We calculated the mean of the difference between the predicted 3D coordinates and the ground truth acceleration (Accel) for the temporal evaluation.

\subsection{Comparison Result and Ablation Study}
% In this section, we evaluate our proposed STR as well as state-of-the-art 3D human pose and shape estimation video methods and single image methods, respectively.
\textbf{Quantitative Comparison.} As shown in Table \ref{tab:tab1}, we first compared our proposed approach with the state-of-the-art video-based and image-based approaches on three datasets. We compare the results with or without the 3DPW dataset (in-the-wild) during training separately. When 3DPW was involved in the training, our method performs admirably on all three test datasets, with the performance on the challenging dataset (3DPW \cite{von2018recovering} and MPI-INF-3DHP \cite{mehta2017monocular}) being particularly impressive. From the quantitative results, we can see that our method provides spatially more accurate 3D human sequence results. In terms of temporal consistency, our approach maintains almost the same acceleration error as TCMR \cite{choi2021beyond}, with only a 0.1\% increase. Our approach improves spatial accuracy while maintaining a similar temporal acceleration error. Our approach focuses on unconstrained scene (extreme lighting, etc.) problems. We also validate on the Human3.6m dataset (indoor), and quantitative results show that under constrained scenarios, our method still outperforms previous methods in spatial accuracy and temporal consistency.
\begin{table*}[htb]
  \centering
    \scalebox{0.58}{\begin{tabular}{cc|cccc|ccc|ccc}
    \toprule
    \multicolumn{2}{c|}{\multirow{2}[4]{*}{Method}} & \multicolumn{4}{c|}{3DPW}     & \multicolumn{3}{c|}{MPI-INF-3DHP} & \multicolumn{3}{c}{Human3.6M} \\
\cmidrule{3-12}    \multicolumn{2}{c|}{} & MPJPE$\downarrow$ & PA-MPJPE$\downarrow$ & MPVPE$\downarrow$ & Accel$\downarrow$ & MPJPE$\downarrow$ & PA-MPJPE$\downarrow$ & Accel$\downarrow$ & MPJPE$\downarrow$ & PA-MPJPE$\downarrow$ & Accel$\downarrow$ \\
\cmidrule{2-12}    \multirow{6}[12]{*}{\rotatebox{90}{single image}} & HMR \cite{kanazawa2018end}  &130.0       &76.7       &-       &37.4       &124.2       &89.8       &-       &88.0       &56.8       &-  \\
\cmidrule{2-12}          & GraphCMR \cite{kolotouros2019convolutional}  &-       &70.2       &-       &-       &-       &-       &-       &-       &50.1       &-  \\
\cmidrule{2-12}          & SPIN \cite{kolotouros2019learning}  & 96.9      &59.2       &116.4       &29.8       &105.2       &67.5       &-       &-       &41.1       &18.3  \\
\cmidrule{2-12}          & I2L-MeshNet \cite{moon2020i2l}  &93.2       & 57.7      &110.1       &30.9       &-       &-       &-       &55.7       &41.1       &13.4  \\
\cmidrule{2-12}          & PyMAF \cite{zhang2021pymaf}  & 92.8      & 58.9      & 110.1      &-       &-       &-       & -      &\bfseries57.7       &40.5       &-  \\
    \midrule
    \multirow{6}[12]{*}{\rotatebox{90}{video}} & HMMR \cite{kanazawa2019learning}     &116.5       & 72.6      & 139.3      & 15.2      &-       &-       &-       & -      &56.9       &-  \\
\cmidrule{2-12}          & Sun ${et}$ ${al.}$ \cite{sun2019human}    & -      &69.5       &-       &-       & -      & -      & -      &59.1      &42.4       &-  \\
\cmidrule{2-12}          & VIBE (\emph{w/o 3DPW})\cite{kocabas2020vibe}     & 93.5      &56.5       &113.4       & 27.1      &97.7       & 63.4      &29.0       & 65.9      & 41.5      & 18.3 \\
\cmidrule{2-12}          &TCMR  (\emph{w/ 3DPW}) \cite{choi2021beyond}     &86.5       & 52.7      & 103.2      & 6.8      &97.6       &63.5       & 8.5      &73.6      &52.0      &3.9 \\

\cmidrule{2-12}          &TCMR  (\emph{w/o 3DPW}) \cite{choi2021beyond}     &95.0       & 55.8      & 111.3 .     & \bfseries6.7      &96.5       &62.8       & 9.5      & 62.3      &41.1      & 5.3 \\

\cmidrule{2-12}          &Lee et al.  (\emph{w/o 3DPW}) \cite{Lee2021UncertaintyAwareHM}   &92.8       & \bfseries52.2      & 106.1      &6.8      &\bfseries93.5       &\bfseries59.4      & 9.4     &58.4     &\bfseries38.4      &6.1 \\

\cmidrule{2-12}          &Ours  (\emph{w/ 3DPW})    &\bfseries85.2       & 52.4      & \bfseries101.2      &6.9      &96.3       &63.1       & 8.6     &73.3     &51.9      &\bfseries3.6 \\

\cmidrule{2-12}          & Ours(\emph{w/o 3DPW})     & 91.5      & 55.2      & 108.7      & \bfseries6.7      & 95.3      & 61.6      & \bfseries8.4      & 67.8      & 46.6      & \bfseries3.6 \\
    \bottomrule
    \end{tabular}} 
    \caption{Comparisons of our approach with state-of-the-art methods on 3DPW(in-the-wild), MPI-INF-3DHP(outdoor), Human3.6M(indoor) testing set. We denote whether 3DPW is involved in the training process as $\emph{w/ 3DPW}$, $\emph{w/o 3DPW}$ respectively.}  \label{tab:tab1}
\end{table*}%
\begin{figure*}[ht]
\centering
\includegraphics[width=0.805\textwidth]{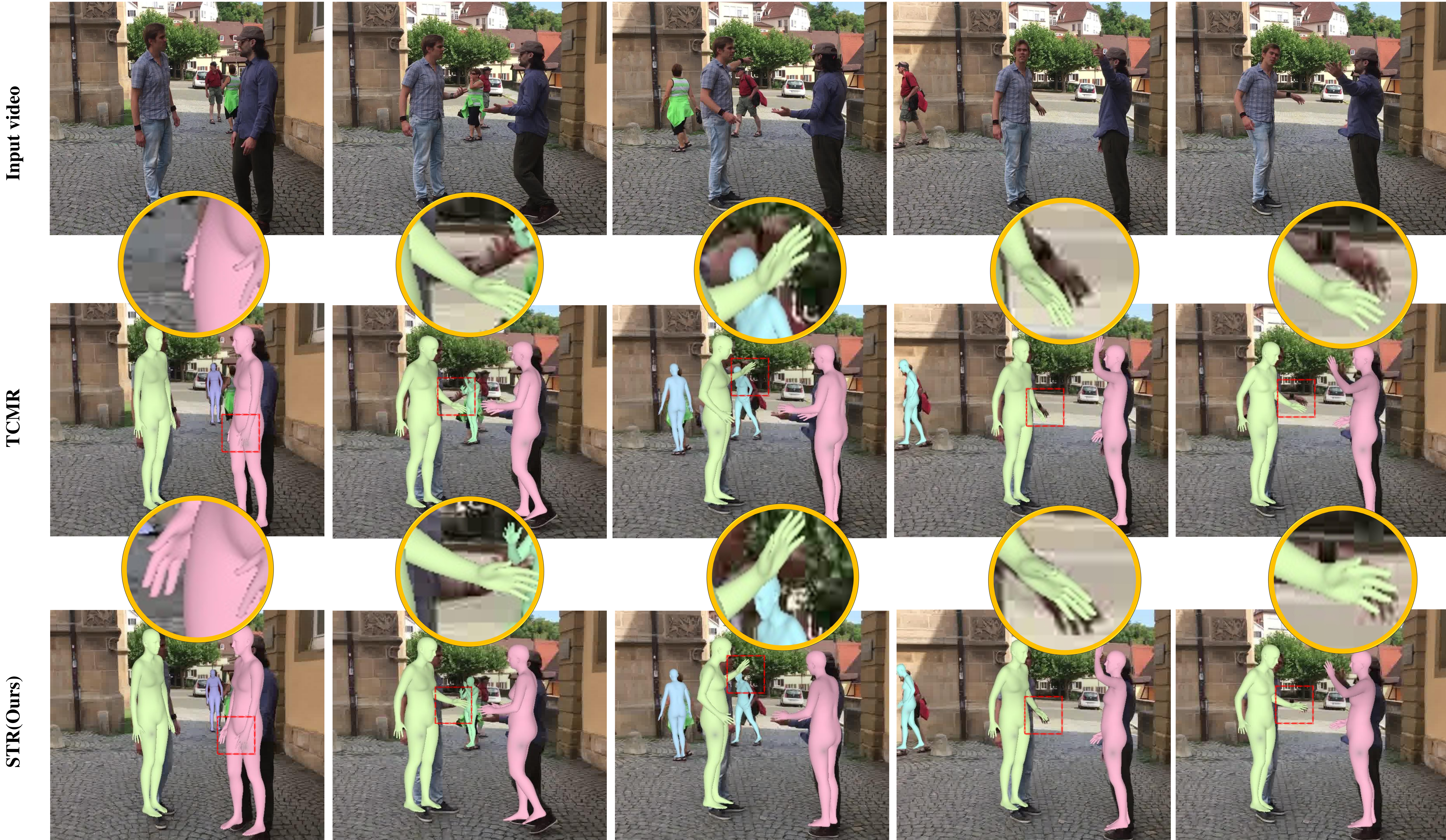}
\caption{Qualitative visualization of STR. The top row shows the original image samples, the middle row shows the TCMR \cite{choi2021beyond} results, and the bottom row shows our results.} \label{fig:fig4}
  
\label{fig8}
\end{figure*}

To verify the effectiveness of our method for unconstrained scenes, our approach is also compared to previous 3D pose and shape estimation algorithms that have not been trained by 3DPW \cite{von2018recovering}. In the in-the-wild dataset 3DPW, our method outperforms TCMR by about 3.5\% and 3.4\% on MPJPE and MPVPE, respectively. It also continues to perform well on MPI-INF-3DHP. When our method is trained without 3DPW, the spatial accuracy and temporal smoothness are still optimal, and the relative improvement is more for ours($w/$ 3DPW training). Our method ensures the plausibility of the human pose by reasoning about the spatio-temporal tendency of human motion. When no wilderness dataset is involved in the training and the constraints become less, our method can still reason and enhance the spatio-temporal tendency in the unconstrained environments and has more robustness to outdoor scenes. 
Notably, Lee et al.'s method \cite{Lee2021UncertaintyAwareHM} generally outperforms our method in accuracy, but weaker than our method in temporal consistency. But since the method of Lee et al. \cite{Lee2021UncertaintyAwareHM} has no published code, we cannot make a qualitative comparison.
Furthermore, the reduction of acceleration errors demonstrates the effectiveness of the proposed spatio-temporal tendency reasoning module. In particular, our method can recover smoother human action sequences compared to single image-based methods, i.e., the temporal consistency is greatly improved. In the indoor dataset, compared with the TCMR($w/o$ 3DPW) \cite{choi2021beyond}, the reason why the MPJPE and PA-MPJPE of Human3.6M \cite{ionescu2013human3} in Table \ref{tab:tab1} is not good in that we have not obtained the SMPL annotations of Human3.6M \cite{ionescu2013human3}.
% As shown in \textbf{Table 1}, we first compared our proposed approach with the most representative video-based methods on the three datasets. As recommended by TCMR \cite{choi2021beyond}, all methods except for HMMR \cite{kanazawa2019learning} are trained on the training set including 3DPW \cite{von2018recovering}, but do not employ Human3.6M \cite{ionescu2013human3} ground truth SMPL parameters by Loper et al \cite{loper2014mosh} for supervision.

\textbf{Qualitative comparison.} In qualitative experiments, as shown in Figures \ref{fig:image1} and \ref{fig:fig4}, our method pays attention to spatial accuracy as well as temporal consistency. TCMR \cite{choi2021beyond} is unable to reason spatial information from more distant frames in the extremely weaker illumination scenes. In addition, because TCMR \cite{choi2021beyond} focuses too much on temporal smoothing enhancement, the human pose variation between frames is relatively small, which also leads to the bias of prediction. In contrast, our method predicts reasonable human action sequences by reasoning the human body information in the current weak illumination from the more distant visible frames. Moreover, our method has a better prediction ability for human movements, especially for limb movements (e.g., walking, arm-waving, etc.). 

\begin{figure*}[ht]
\centering
\includegraphics[width=0.9\textwidth]{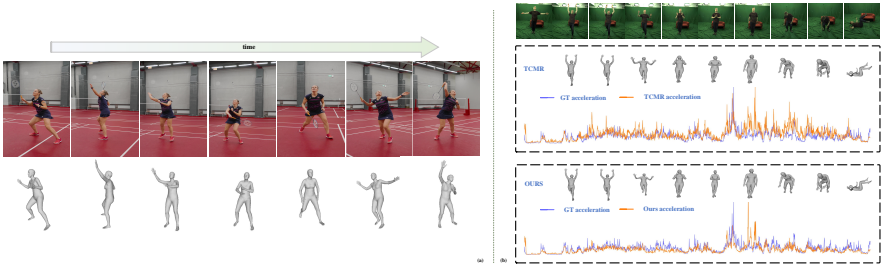}
\caption{Subfigure (a) is our reconstructed video sequence from the web. Subfigure (b) is the comparison among TCMR, Ours, and GT acceleration errors.}  \label{fig:fig5}
\label{fig1}
\end{figure*}

\textbf{Discussion.} Figure \ref{fig:fig5}(a) shows our reconstructed out-of-dataset video sequence. Our method can predict human actions with a continuous tendency in consecutive frames and has promising generalization capabilities. The transitional properties of the actions show that our method captures the past spatio-temporal tendency and predicts the future spatio-temporal tendency. As shown in Figure \ref{fig:fig5}(b), compared to TCMR \cite{choi2021beyond}, our acceleration error curves are generally flat and have similar trends. As the time step increases, our acceleration error approaches the GT acceleration error. At most time steps, our acceleration error is even lower than the GT acceleration error, which indicates that our method reasons for the correct spatio-temporal tendency of human motion which validates the effectiveness of our method.

\textbf{Ablation Study.} Table \ref{tab:tab2} shows that the acceleration error rises by 0.3 and the accuracy decreases by 0.1 after removing the TTR module. When we remove the STE module and the time-frequency domain enhancement module in STE, respectively, the PA-MPJPE increases. This indicates that the model cannot perceive the time-domain sensitive or frequency-domain high-frequency human spatial motion tendency, resulting in a decrease in accuracy. We removed the self- and cross-integration strategy from the integration strategies module and the PA-MPJPE and acceleration errors rise. This shows that the inter-frame features need to complement each other to refine the current spatio-temporal features to recover reasonable human poses. 

\begin{table}[ht]

\centering
\setlength{\tabcolsep}{10mm}{
\renewcommand{\arraystretch}{1.3}
\scalebox{0.7}{\begin{tabular}{c|c|cc} 
\specialrule{0.125em}{4pt}{0.1pt}
  \textbf{Model}  & \textbf{\textit{PA-MPJPE}}$\downarrow$& \textbf{\textit{Accel}}$\downarrow$ \\
  \hline  
%   \specialrule{0.08em}{1pt}{3pt}
 
%  \specialrule{0.08em}{1pt}{1pt} 
%   IoU$\uparrow$      &0.605& 0.607&0.608& \textbf{0.619} \\

    STR w/o TTR       & 61.9 & 8.7   \\
    STR w/o STE              &  62.1 & \bfseries8.4 \\
    STR w/o STE (time-domin)              &  61.9 & \bfseries8.4 \\
    STR w/o STE (frequency-domin)            &  61.7 & 8.5 \\
    STR w/o Integration strategies(self-)             & 62.3 & 9.1 \\
    STR w/o Integration strategies(cross-)             & 62.7 & 9.0 \\
    STR                        &  \bfseries61.6 & \bfseries8.4 \\
  \specialrule{0.125em}{0.1pt}{0.1pt}
\end{tabular}}
}
% \vspace{0.2cm}
\caption{Effects of the network designs on the performance on the MPI-INF-3DHP dataset.}\label{tab:tab2}
\end{table}

\section{Conclusion}
\label{sec:Conclusion}
We offer a spatio-temporal tendency reasoning method for human pose and shape estimation from videos. STR fully exploits human feature representations in video sequences and enhances spatio-temporal feature representations by reasoning about temporal information representations and exciting sensitive spatial features in human motion sequences. Spatio-temporal features are also refined through integration strategies. We demonstrate that our method provides smooth and accurate human motion through extensive testing.

\section{Acknowledgements}
This work was supported by the National Natural Science Foundation of China under Grant 62062056, in part by the Ningxia Graduate Education and Teaching Reform Research and Practice Project 2021, and in part by the National Natural Science Foundation of China under Grant 61662059.

\bibliography{egbib}
\end{document}